\documentclass[pdflatex,sn-mathphys-num]{sn-jnl}

\usepackage{graphicx}%
\usepackage{multirow}%
\usepackage{amsmath,amssymb,amsfonts}%
\usepackage{amsthm}%
\usepackage{mathrsfs}%
\usepackage[title]{appendix}%
\usepackage{xcolor}%
\usepackage{textcomp}%
\usepackage{manyfoot}%
\usepackage{booktabs}%
\usepackage{algorithm}%
\usepackage{algorithmicx}%

\usepackage{algpseudocode}%
\usepackage{listings}%
\usepackage{bm}%
\usepackage{gensymb}%
\usepackage{placeins}
\usepackage{float}
\usepackage{lineno}

\usepackage[nomarkers, nolists]{endfloat}

\usepackage{pdflscape}
\usepackage{xcolor}
\definecolor{mydarkgreen}{RGB}{0,139,0}
\theoremstyle{thmstyleone}

\theoremstyle{thmstyletwo}%

\theoremstyle{thmstylethree}%
\begin{document}

\title[Article Title]{{Surface-Based Manipulation with Modular Foldable Robots}}

\author[1]{\fnm{Ziqiao} \sur{Wang}}\email{ziqiao.wang@epfl.ch}
\equalcont{These authors contributed equally to this work.}

\author[1]{\fnm{Serhat} \sur{Demirtas}}\email{serhat.demirtas@epfl.ch}
\equalcont{These authors contributed equally to this work.}

\author[1]{\fnm{Fabio} \sur{Zuliani}}\email{fabio.zuliani@epfl.ch}

\author*[1]{\fnm{Jamie} \sur{Paik}}\email{jamie.paik@epfl.ch}

\affil[1]{\orgdiv{Swiss Federal Institute of Technology Lausanne (EPFL)}, \orgname{Reconfigurable Robotics Laboratory}, \orgaddress{\city{Lausanne}, \postcode{1015}, \country{Switzerland}}}

\abstract{Intelligence lies not only in the brain (decision-making processes) but in the body (physical morphology). The morphology of robots can significantly influence how they interact with the physical world, crucial for manipulating objects in real-life scenarios. Conventional robotic manipulation strategies mainly rely on finger-shaped end effectors. However, achieving stable grasps on fragile, deformable, irregularly shaped, or slippery objects is challenging due to difficulty in establishing stable forces or geometric constraints. 

Here, we present surface-based manipulation strategies that diverge from classical grasping approaches, using flat surfaces as minimalist end-effectors. By adjusting surfaces' position and orientation, objects can be translated, rotated, and flipped across the surface using closed-loop control strategies. Since this method does not rely on stable grasping, it can adapt to objects of various shapes, sizes, and stiffness levels, even enabling manipulation of the shape of deformable objects. Our results provide a new perspective for solving complex manipulation problems.}

%TC:endignore

\maketitle
\newpage

\section{Introduction}\label{sec1}

\begin{figure*}[ht!]
  \centering
  \includegraphics[height=0.81\textheight]{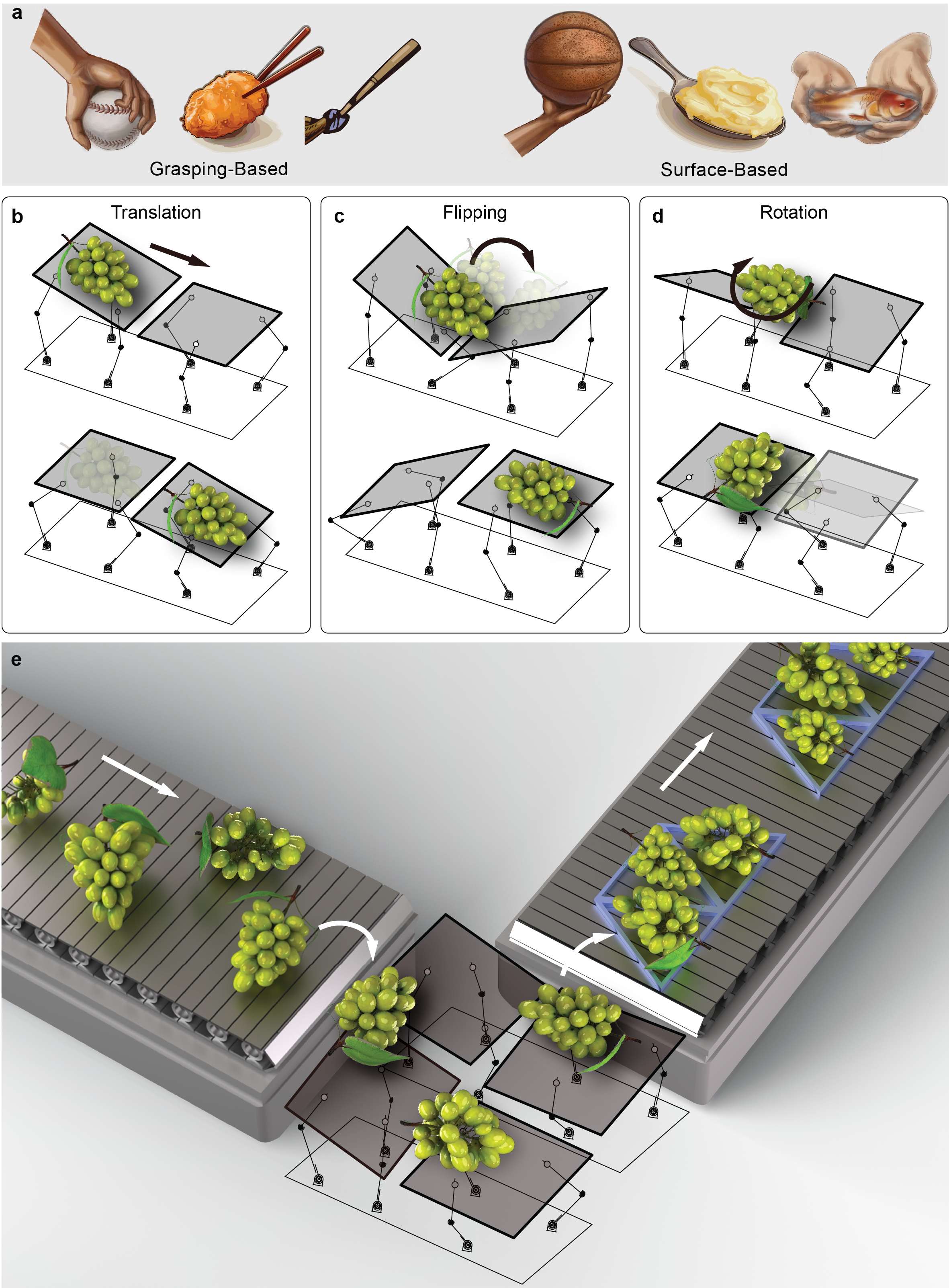}
  \caption{\textbf{Surface-based manipulation.} (\textbf{a}) Object properties influence whether we intuitively grasp or support them with a surface. Small, rigid objects with defined edges are easily grasped, while large, round, deformable, soft, or slippery objects are more effectively manipulated through surface contact. This distinction highlights how surface-based manipulation complements traditional grasping in handling diverse objects. Building upon this, we present a novel approach to robotic object manipulation using surfaces. This is achieved through the integration of three main motion principles: (\textbf{b}) translation, (\textbf{c}) flipping, and (\textbf{d}) rotation. (\textbf{e}) An example is in food packaging automation, where it addresses the challenge of handling items with varied shapes without causing damage. Surface-based manipulation enables tasks like rotating or positioning grape bunches for packaging and inspection.}
  \label{fig:first}
\end{figure*}

Physical morphologies significantly shape how a system interacts with and perceives its environment \cite{embodiment, sun2023embedded,baines2024robots,shah2021soft,baines2022multi,gupta2021embodied}. Morphologies that are structurally adapted to a task reduce the complexity of control and enable more precise and responsive task execution. In nature, many animals have evolved diverse grasping mechanisms to adapt to various contexts—for example, manual grippers resembling anthropomorphic hands \cite{pouydebat2023convergent}, spinal grippers such as the prehensile tails of monkeys and certain lizards, and muscular hydrostats like octopus tentacles and elephant trunks \cite{softgrip}. Roboticists draw inspiration from the biological systems to develop grippers that utilize mechanical interlocking for geometric constraints \cite{katzschmann,hao_2023,howard_2022}, friction for force constraints \cite{7390277,9000610,10146043,7994658}, or adhesion for stabilization \cite{Chen_2021,shintake_2016,piskarev_2023}. These methods form the basis of most current robotic gripper designs \cite{ZHANG2020105694}. However, they often face challenges when manipulating irregularly shaped, deformable, or fragile objects \cite{doi:10.1126/science.aat8414}. Purely geometric constraints struggle with irregular shapes, and force constraints can damage deformable or delicate items. To address these challenges, researchers have developed advanced gripper designs that enhance robotic manipulation. Inspired by biological systems, soft robotic grippers with mechanical compliance can handle fragile objects \cite{SoftActuators1, softgriperreview}. Granular jamming grippers conform to objects of arbitrary shape to enable their pick-and-place operations \cite{GranularJamming1, GranularJamming2}, while tactile sensing improves real-time adjustments for delicate objects \cite{tactilesensing, GranularJamming3}. These methods still depend on stable grasps that match the size of the object, and struggle with deformable, flat, or highly variable objects \cite{softgriperreview}. Reflecting on our daily experiences (see Fig. \ref{fig:first}a), we often address similar challenges by adopting surface-based, nonprehensile manipulation methods and providing support from below. For example, while chopsticks are effective for picking up solid foods like fried potatoes, a spoon is more suitable for deformable substances like mashed potatoes. We hold a basketball with our palms rather than gripping it solely with our fingers, and we support a slippery fish from underneath. These examples demonstrate that when traditional grasping is ineffective, we naturally resort to surface-based, nonprehensile manipulation strategies. Motivated by this observation, introducing surface-based end effectors into robotics by adapting their morphology to planar forms offers a promising solution to these challenges.

Robotics research has introduced devices capable of dynamically altering their surface geometries through actuation mechanisms \cite{rasmussen2012shape,alexander2018grand,follmer2013inform,gronbaek2020kirigamitable,steed2021mechatronic,hwang2022shape,tahouni2020nurbsforms,siu2018shapeshift}. By adjusting their configurations, these systems can adapt to and interact more effectively with various objects. One approach employs grids of individually actuated parallel robots or linear actuators to create distributed manipulation surfaces \cite{Thompson2021, patil2023, follmer2013inform, xue2023arraybot}. Another utilizes origami-inspired structures integrated into robotic gripper surfaces to modulate contact friction \cite{Lu2020c}. To enhance the capabilities of reconfigurable surfaces, modular designs incorporating soft actuators powered by vacuum \cite{Robertson2018} or origami-inspired actuators \cite{Salerno2020} have been explored. Hu et al. used pneumatic morphological transformation to adjust wettability and manipulate droplets on a surface undergoing deformation \cite{hu2022}. Surface-based manipulation techniques have also been applied at smaller scales, specifically in the reconfigurable braille displays. These displays use dielectric elastomers \cite{Qu2021}, electromagnetism \cite{Bettelani2020}, or shape-memory alloys \cite{Haga2005}. Furthermore, reconfigurable surfaces find applications in shape-changing displays designed for human interaction \cite{alexander2018grand, steed2021mechatronic, gronbaek2020kirigamitable, siu2018shapeshift,stanley2016closed}. An alternative approach to address complex manipulation tasks involves integrating nonprehensile manipulation strategies that impose unilateral constraints on the object \cite{ruggiero_2018}. For instance, \cite{R2cite2} demonstrates a paddle-like end-effector with visual feedback for pick-and-place tasks on rigid hexahedral objects, while \cite{R2cite3} employs two 6 degrees of freedom (DoF) robotic arms with dual “palms” for reorientation tasks. In \cite{R2cite1, R2cite6}, a single motor-driven flexible joint mechanism enables planar manipulation via vibration modes. Inspired by pizza paddles, \cite{R2cite4, R2cite5} introduce planar end-effectors facilitating object translation and rotation, and a related method uses two serial manipulators for rapid, coordinated actions like stir-frying \cite{liu_2022}. {While some of these systems incorporate visual feedback \cite{R2cite2, R2cite1} or explore deformable object manipulation \cite{R2cite5, higashimori2013gait}, many lack closed-loop control, focus on limited range of rigid objects, or rely on single-surface methods that limit control over contact forces and scalability.}

Modular reconfigurable surface-based strategies also face significant practical limitations. While the potential has been demonstrated in manipulating spheres \cite{liu2021robotic, johnson2023multifunctional}, challenges such as limited workspace and slow deformation speeds pose an obstacle for broader scenarios. Additionally, the necessity for continuous surface deformation reduces force output and limits their utility in manipulation contexts. These strategies frequently require the coordinated operation of many modules to manipulate a single object, complicating scalability to larger module sizes and quantities. Consequently, current applications lack dexterous strategies for handling non-rigid or irregularly shaped items and are often confined to specific hardware platforms. These limitations emphasize the need for novel design and actuation methodologies. Exploring innovative approaches to reconfigurable surfaces could significantly advance robotic manipulation, expanding robots' capabilities in handling a diverse range of objects. 

Current nonprehensile manipulation methods on reconfigurable surfaces, which utilize individually actuated units, rely heavily on the kinematics of the modules and the sizes of the objects being manipulated. Together, these factors determine the types of nonprehensile motion primitives that can be utilized. Consequently, developing a comprehensive framework to analyze and compare these platforms is crucial for optimizing their design and manipulation strategies. Using information obtained from published literature on several platforms \cite{wavehandling, Robertson2018, follmer2013inform, xue2023arraybot, R2cite2, R2cite3, R2cite1, R2cite4, omniawheel, platemanipulator, patil2023, softmachinetable, rodyman}, we developed a scaling model for surface-based manipulation that results in the mapping of mobility and manipulability of these platforms. Finding the optimal setup for surface-based manipulation involves determining the exact number of DoF needed for each module and how many modules are necessary for different manipulation tasks. Table \ref{sp:scalability} lists various surface-based object manipulation examples and summarizes their capabilities in terms of repositioning and reorienting objects placed on them. We standardize the module dimensions across the platforms to facilitate comparative analysis, study the relationship between the working modules and manipulability, and highlight their capability in accommodating objects of diverse sizes. The minimum manipulation workspace, {$MW_{min}$}, indicates the smallest area where both the functions of the reconfigurable surface and the manipulation strategies are effectively employed. Then, we calculate the minimum mobility requirement, \textbf{$MR_{min}$}, that represents the necessary number of actuators for different types of surface-based manipulation techniques using single unit DoF, \textbf{$DoF_{SU}$}, and surface size, \textbf{$SS_{SU}$}: 

\begin{equation}
\label{eq:mobility}
MR_{min}
=\frac{DoF_{SU}\times MW_{min}}{SS_{SU}}.
\end{equation}

\begin{table}[h!]

\centering
\caption{\textbf{Overview of existing surface-based manipulation strategies.}}
\label{sp:scalability}
\setlength{\tabcolsep}{1.2pt}
\begin{tabular}{ccccccccc}
\hline
\multirow{4}{*}{Platform} & \multirow{4}{*}{\begin{tabular}[c]{@{}c@{}}Single Unit\\ DoF,\\ $DoF_{SU}$\end{tabular}} & \multicolumn{3}{c}{\multirow{3}{*}{Manip. Modes}} & \multirow{4}{*}{\begin{tabular}[c]{@{}c@{}}Single Unit\\ Surface Size,\\ $SS_{SU}$\end{tabular}} & \multirow{4}{*}{\begin{tabular}[c]{@{}c@{}}Object\\ Size\end{tabular}} & \multirow{4}{*}{\begin{tabular}[c]{@{}c@{}} Minimum\\ Manip.\\ Workspace,\\ $MW_{min}$\end{tabular}} & \multirow{4}{*}{\begin{tabular}[c]{@{}c@{}}Minimum\\ Mobility \\ Requirement,\\ $MR_{min}$\end{tabular}} \\
 &  & \multicolumn{3}{c}{} &  &  &  &  \\
 &  & \multicolumn{3}{c}{} &  &  &  &  \\
 &  & \multicolumn{1}{l}{Tra.} & \multicolumn{1}{l}{Rot.} & \multicolumn{1}{l}{Flip.} &  &  &  &  \\ \hline
Single Act. \cite{R2cite1} & 1 & \checkmark &  \checkmark & & $d \times d$ & \textless $d \times d$ & $d\times d$ & 1  \\
WaveHandling \cite{wavehandling} & 1 & \checkmark &  & \multicolumn{1}{c}{} & $d \times d$ & \textgreater $d \times d$ & $2d \times 2d$ & 4  \\
Soft Surface \cite{Robertson2018} & 1 & \checkmark &  & \checkmark & $d\times d$ & \textgreater $2d\times2d$ & $3d\times3d$ & 9 \\
inFORM \cite{follmer2013inform} & 1 & \checkmark & \checkmark & \checkmark & $d\times d$ & \textgreater $2d\times2d$ & $3d\times3d$ & 9 \\
ArrayBot \cite{xue2023arraybot} & 1 & \checkmark & \checkmark & \checkmark & $d\times d$ & \textgreater $2d\times2d$ & $3d\times3d$ & 9 \\
Pizza Peel \cite{R2cite4} & 2 & \checkmark & \checkmark & & $d\times d$ & \textless $d\times d$ & $d\times d$ & 2 \\ 
Omnia Wheel \cite{omniawheel} & 3 & \checkmark & \checkmark & \multicolumn{1}{c}{} & $d\times d$ & \textgreater $2d\times2d$ & $3d\times d$ & 9 \\
Planar Manip. \cite{platemanipulator} & 3 & \checkmark & \checkmark & \multicolumn{1}{c}{} & $d\times d$ & \textless $d\times d$ & $d\times d$ & 3 \\
Delta Arrays \cite{patil2023} & 3 & \checkmark & \checkmark & \checkmark & $d\times d$ & \textgreater $2d\times2d$ & $3d\times3d$ & 27 \\
Soft Table \cite{softmachinetable} & 4 & \checkmark & \checkmark & \multicolumn{1}{c}{} & $d\times d$ & \textgreater $2d\times2d$ & $3d\times3d$ & 36 \\
Dynamic Manip. \cite{R2cite2} & 6 & \checkmark & \checkmark & \checkmark & $d\times d$ & \textless $d\times d$ & $d\times d$ & 6 \\ 
Two-Palm \cite{R2cite3} &6 & \checkmark & \checkmark & \checkmark & $d\times d$ & \textless $d\times d$ & 2$d\times d$ &12 \\ 
RoDyMan \cite{rodyman} & 12 & \checkmark & \checkmark & \multicolumn{1}{c}{} & $d\times d$ & \textless $d\times d$ & $d\times d$ & 12 \\ 
\hline
\hline
\textbf{This Work} & \textbf{3} & \textbf{\checkmark} & \textbf{\checkmark} & \textbf{\checkmark} & $\bm{d\times d}$ & $\bm{<d\times d}$ & $\bm{2d\times d}$ & \textbf{6} \\ \hline
\end{tabular}
\end{table}

Surfaces consisting of multiple units with one $DoF_{SU}$ can form different profiles from a flat surface, such as slopes for rolling objects or ridges for flipping them, and they enable manipulation of objects larger than $SS_{SU}$ \cite{follmer2013inform, Robertson2018, xue2023arraybot, wavehandling}. In these examples, the $MR_{min}$ and consequently the actuator number typically remain relatively high, up to 900, mainly due to the necessary motion primitives and the properties of the manipulated objects \cite{follmer2013inform, Robertson2018, xue2023arraybot, wavehandling}. Increasing the $DoF_{SU}$ enables the utilization of additional motion primitives with fewer units. For instance, an example system comprises three $DoF_{SU}$ arranged in an 8×8 configuration \cite{patil2023, Thompson2021}. These units, having finger-like end effectors, manipulate objects through coordinated movements such as sliding in addition to rolling \cite{patil2023, Thompson2021}. Similarly, Deng et al. introduced a pneumatically actuated soft modular surface with four $DoF_{SU}$ capable of translating and rotating objects through small deformations on the surface \cite{softmachinetable} and complementary trajectory planning \cite{softtable_traj}. Although the number of units is lower in examples with higher $DoF_{SU}$, the $MR_{min}$ remains high due to the increased actuator number requirements. The introduction of dynamic movements such as vibration has also been shown to facilitate both translation and rotation within the plane \cite{platemanipulator}. Utilizing a similar methodology, surface-based manipulation extends to serial manipulators through the use of flat-surface end effectors. Ruggiero et al. utilized a dual-arm robot with a total of 12 DoF and employed dynamic manipulation techniques such as rolling, tossing, and batting that translate or rotate objects \cite{ruggiero_2018}. Analyzing the $MR_{min}$ aspects of the state-of-the-art reveals a trade-off between the number of required modules and the $DoF_{SU}$. For instance, a higher $DoF_{SU}$ for fewer units to accomplish the same task. However, increasing the $DoF_{SU}$ poses challenges due to hardware limitations, making it more difficult to expand the $SS_{SU}$ by adding more modules. Consequently, we adopt two units with three $DoF_{SU}$ that represent the minimum configuration required to construct a reconfigurable surface capable of performing dexterous manipulation tasks. This surface consists of individual elements arranged in flat configuration that is able to form three-dimensional configurations and provide synchronized movements upon actuation. The three $DoF_{SU}$, along with the utilization of the interaction between the two modules, enable a range of dynamic manipulation techniques. {This configuration ensures the lowest $MR_{min}$ for surface-based manipulation platforms that can perform object translation, rotation, and flipping (see Fig. \ref{fig:first}b–d). While some prior work \cite{R2cite2} achieves a similar $MR_{min}$, using a single platform with six DoF to perform comparable manipulation modes, the modularity in our system distributes the required MR across two smaller units, each with three DoF. This bimanual design enhances dexterity by enabling the simultaneous application of forces from multiple directions, allowing for more dexterous tasks such as folding deformable objects. It also improves scalability by enabling incremental expansion of functionality through additional modules, rather than increasing complexity within a single unit. Moreover, our system supports closed-loop control and quantitative evaluation and handles diverse forms of objects. The system exhibits scalability in terms of module size, module number, and the size of the object being manipulated. Depending on the specific requirements of the objects, the proposed strategies could be adopted to accommodate different module sizes. Furthermore, increasing the number of modules allows for the application of these strategies to multiple objects simultaneously.}

We propose manipulation strategies based on reconfigurable modular surfaces to perform fundamental manipulation tasks such as translating, rotating, and flipping objects of various forms, including non-rigid and irregularly shaped items. We demonstrate these strategies using two identical modular and foldable robots placed side by side to form a reconfigurable surface operating through a decoupled actuation procedure. Each robot is designed, manufactured, and assembled using an origami-based approach combined with the Canfield joint principle \cite{Canfield_1998,Salerno2020, mintchev2019portable, Mete2021, robertson2021soft} and has three DoF. The hierarchical closed-loop control architecture overcomes kinematic constraints and adapts to different robotic platforms, which addresses hardware adaptability issues. By eliminating the need for grasping and continuous surface deformation, our method enables the manipulation of objects with diverse shapes and forms. Quantitative closed-loop experiments validate the overall performance of our system and demonstrate that our approach achieves more dexterous tasks with fewer DoF compared to other surface-based methods. This work also shows the potential of combining different manipulation strategies to alter the shape of flexible and deformable objects, overcoming existing limitations and expanding the capabilities of robotic manipulation systems. Our approach offers a new perspective for applications such as food packaging (see Fig. \ref{fig:first}e) involving reconfigurable surfaces, where automation faces the challenge of handling items with varied and unpredictable shapes without causing damage.

\section{Results}\label{sec3}

\begin{figure*}[t]
     \centering
     \includegraphics[width=\textwidth]{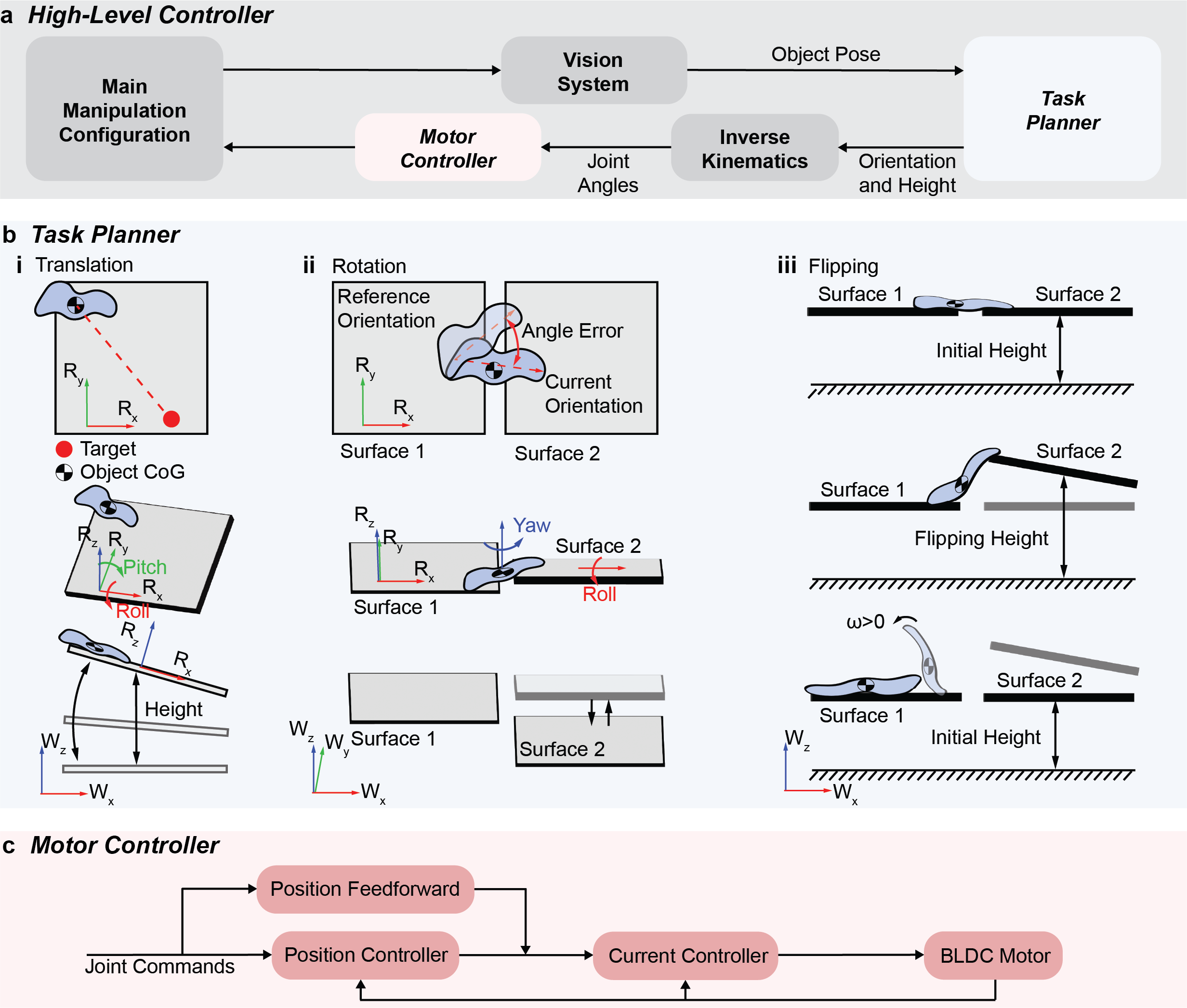}
     \caption{\textbf{Control schematic for surface-based object manipulation.} (\textbf{a}) High-level controller involves five main components. Main manipulation configuration includes two modules side by side. The vision system captures the object’s position and orientation and provides real-time feedback to task planner. (\textbf{b}) The task planner, a state machine, defines surface orientations and height and then these are converted into joint reference positions via inverse kinematics. It transitions among three tasks: (\textbf{i}) translation, (\textbf{ii}) flipping, (\textbf{iii}) and rotation. Translation adjusts the surface's roll and pitch to move the object, followed by height oscillations. Rotation has two phases: Phase 1 adjusts Surface 2's roll, and Phase 2 raises and lowers it to repeat Phase 1 until the desired angle is achieved. Flipping occurs through the rapid elevation of Surface 2. (\textbf{c}) The low-level controller uses BLDC motors with a servo position controller to track joint angle commands with real-time feedback.}
     \label{fig:control_scheme}
\end{figure*}

\begin{figure}[t]
     \centering
     \includegraphics[width=
     \textwidth]{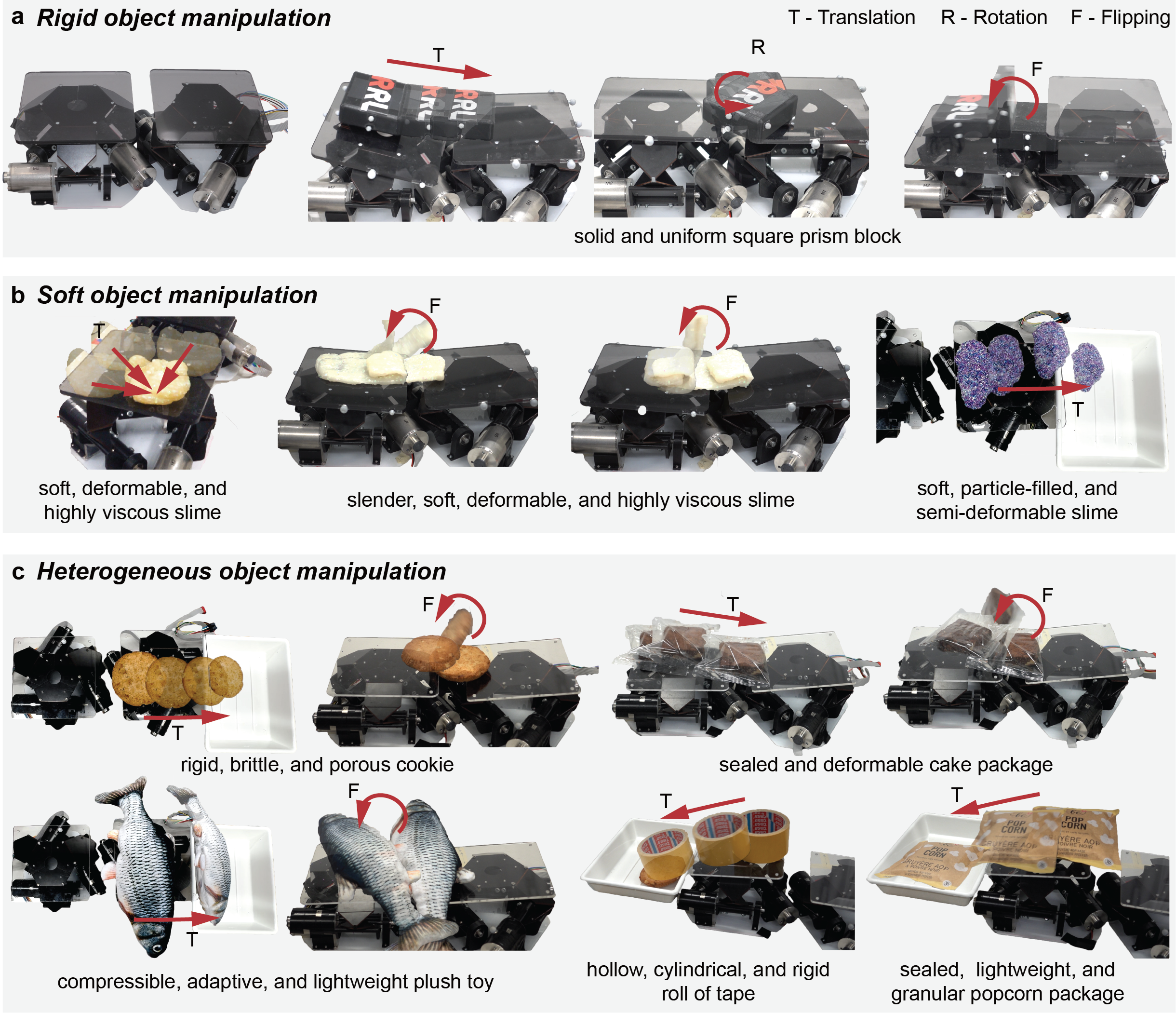}
     \caption{\textbf{Surface-based manipulation of various types of objects.} A rigid object (a solid, uniform square prism block) and the first two white soft objects (a malleable slime) are manipulated using closed-loop control, while the remaining objects are manipulated via open-loop teleoperation to transition between different states, demonstrating the strategy’s versatility across diverse object types. (a) Rigid object transfer (T), rotation (R), and flipping (F). (b) Manipulation of soft and deformable objects: adjusting the position or shape of deformable materials in various forms. (c) Manipulation of different objects, from top left to bottom right: a cookie, a cake in a transparent package, a cotton-stuffed toy fish, a roll of tape, and a bag of unevenly distributed popcorn.}
     \label{fig:movie}
\end{figure}

The surface-based manipulation in this work involves three strategies: translation, rotation, and flipping (see Fig. \ref{fig:control_scheme}). To demonstrate the generalizability of surface-based manipulation, we first conducted open-loop teleoperated experiments on a variety of objects and substances of different shapes, sizes, and mechanical properties, including packaged cookies, cakes, a cotton-stuffed toy fish, a roll of tape, and a bag of popcorn (Fig. \ref{fig:movie} and Supplementary Movie 1). We achieved consistent and successful manipulation as our surface-based approach inherently avoids the complexities of grasping, it can accommodate diverse shapes while minimizing stress on the objects.

Building upon the initial demonstrations, we conducted a series of quantitative, vision-based, closed-loop experiments with two modules to evaluate the proposed strategies. Incorporating perception capabilities, our closed-loop control algorithms enhanced the precision and robustness of these manipulation tasks. First, we validated the translation task, an essential foundation for several subsequent tasks, by repositioning objects to desired locations on the entire surface (Fig. \ref{fig:result_of_softtransfer} and Supplementary Movie 2). Next, we combined translation and flipping strategies to fully flip planar objects, effectively swapping their top and bottom surfaces, as shown in Fig. \ref{fig:result_of_rotation}a-b and Supplementary Movie 3. The third experiment involves changing the orientation of the object in the middle of two surfaces (Fig. \ref{fig:result_of_rotation}c-d and Supplementary Movie 4). Beyond these pick-and-place tasks, we demonstrated more dexterous manipulations by reshaping a deformable object by folding it to alter its length, as shown in Fig. \ref{fig:result_of_softf} and Supplementary Movie 5. 

\subsection{Translation of Soft and Randomly Shaped Objects}
Translating an object is essential in robotic manipulation for repositioning objects and preparing them for subsequent operations. In this work, the first translation analysis involved a randomly shaped object, Play-Doh putty, specifically chosen for its soft and deformable properties. These characteristics present a challenge for conventional grippers, making the object an ideal candidate for testing the capability of our strategy in handling complex material properties.

\begin{figure}[t!]
     \centering
     \includegraphics[width=0.65\textwidth]{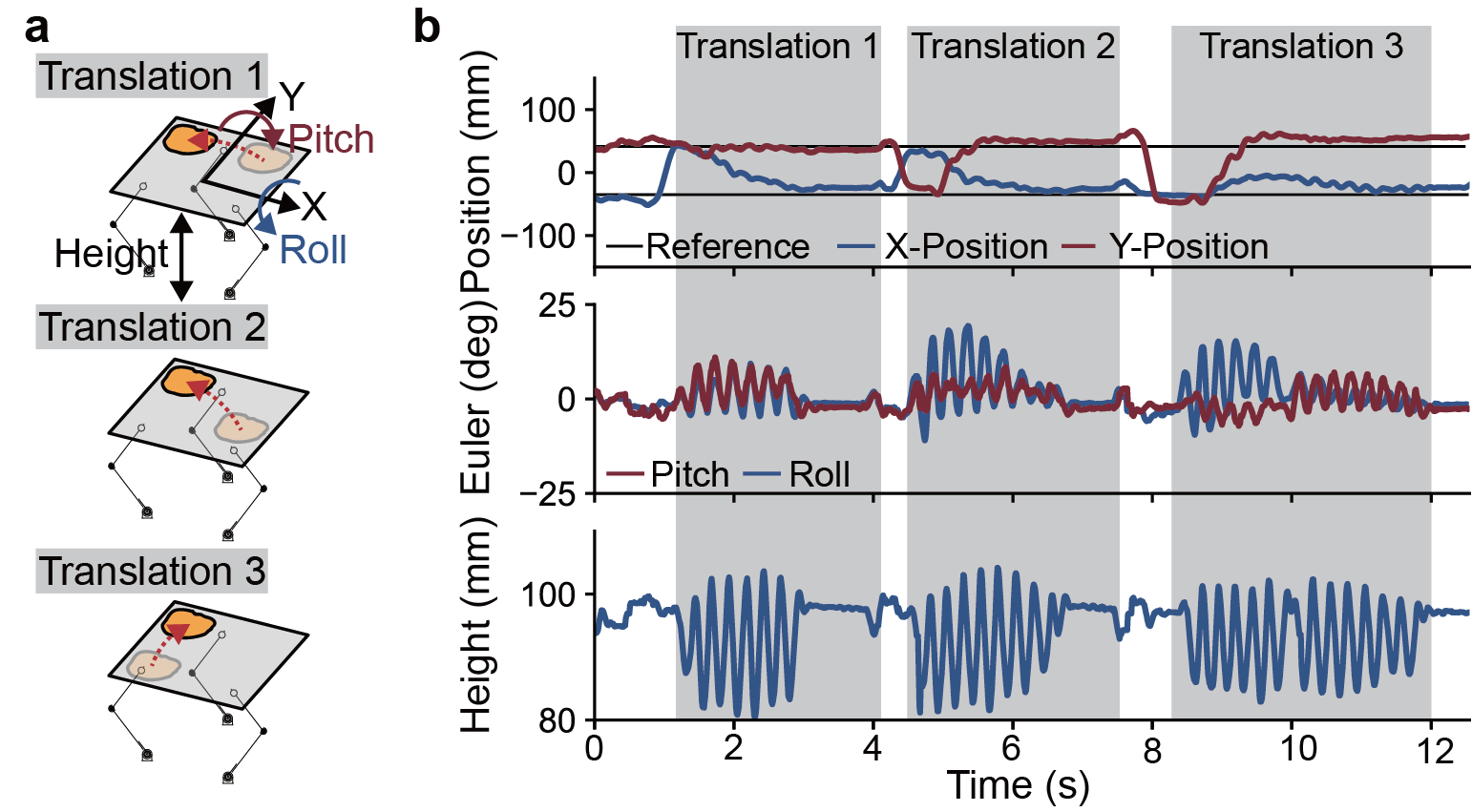}
     \caption{
     \textbf{Translation of soft objects.} In this experiment, the efficacy of our translation strategy was assessed using a deformable object positioned at multiple initial locations. (\textbf{a}) Initial setup for three sequential tests with object: Lighter colors indicate initial positions and darker colors show target positions. Objects are manually reset to new initial positions after each translation. (\textbf{b}) The object's trajectory in the XY-plane and surface adjustments in roll, pitch, and height are controlled by the transfer strategy.}
     \label{fig:result_of_softtransfer}
\end{figure}

The objective is to move the soft object across the surface, starting from varying initial locations but converging to a consistent reference position by closed-loop control (see Fig. \ref{fig:movie}b and Fig. \ref{fig:result_of_softtransfer}). In Fig. \ref{fig:result_of_softtransfer}a, we illustrate three sequential manipulation paths, each initiated from a distinct starting position. Upon reaching the target position and achieving a steady state, we manually repositioned the object to a new starting point. Despite the different starting points, each route consistently concluded at the same predetermined location in the top-left quadrant of the surface. As shown Fig. \ref{fig:result_of_softtransfer}b, the object consistently returns to its target after manual displacements, guided by the control strategy and continuous adjustments in surface orientation and height. During Translation 1, positional changes along the X-axis induced significant variations in surface’s pitch. Before Translation 2, placing the object diagonally opposite the target influenced both the surface’s pitch and roll, with roll adjustments more pronounced due to asymmetries in the workspace (see Supplementary Fig. 1 and Supplementary 2). In Translation 3, the primary deviation along the Y-axis mainly affected surface’s roll values, while minor X-axis changes near the end introduced slight pitch alterations. Additionally, height adjustments in each control cycle caused oscillations that moved the object toward its target.

Across six trials (see Supplementary Movie 2), the object achieved average velocities of 5.62 cm/s along X-axis and 5.55 cm/s along Y-axis, with steady-state errors of 1.14 cm and 0.87 cm, respectively. Subtle positional discrepancies likely arose from differences in achievable tilt angles (Supplementary Fig. 2). Overall, these results confirm that our manipulation strategy reliably directs objects to their intended targets, even from various starting points, while minimizing deformation of soft objects.

%TC:ignore

%TC:endignore

\subsection{Translation and Flipping of Rigid Object}
Flipping motions are crucial in both industrial applications such as assembly and food inspection on production lines \cite{10.1007/978-3-031-13841-6_44} and domestic robotics for household tasks \cite{10163971}. Flipping typically requires multiple controllable contact points \cite{kuffner2016motion} and necessitates complex manipulators or dual-arm robots working in coordination \cite{kim2021integrated}. It also demands accurate modeling of contact interactions, handling uncertainties, and maintaining grasp stability. Additionally, limitations in the end effector's workspace and kinematics further complicate the execution of flipping motions. These complexities make traditional manipulation strategies inadequate for efficiently automating flipping tasks. Our approach addresses these issues by removing workspace dependence through nonprehensile dynamic motion enabled by coordinated modular interactions. This strategy executes complex tasks like 180$\degree$ flipping along a horizontal axis without intricate grasping mechanisms or high DoF.

\begin{figure*}[t]
     \centering
     \includegraphics[width=\textwidth]{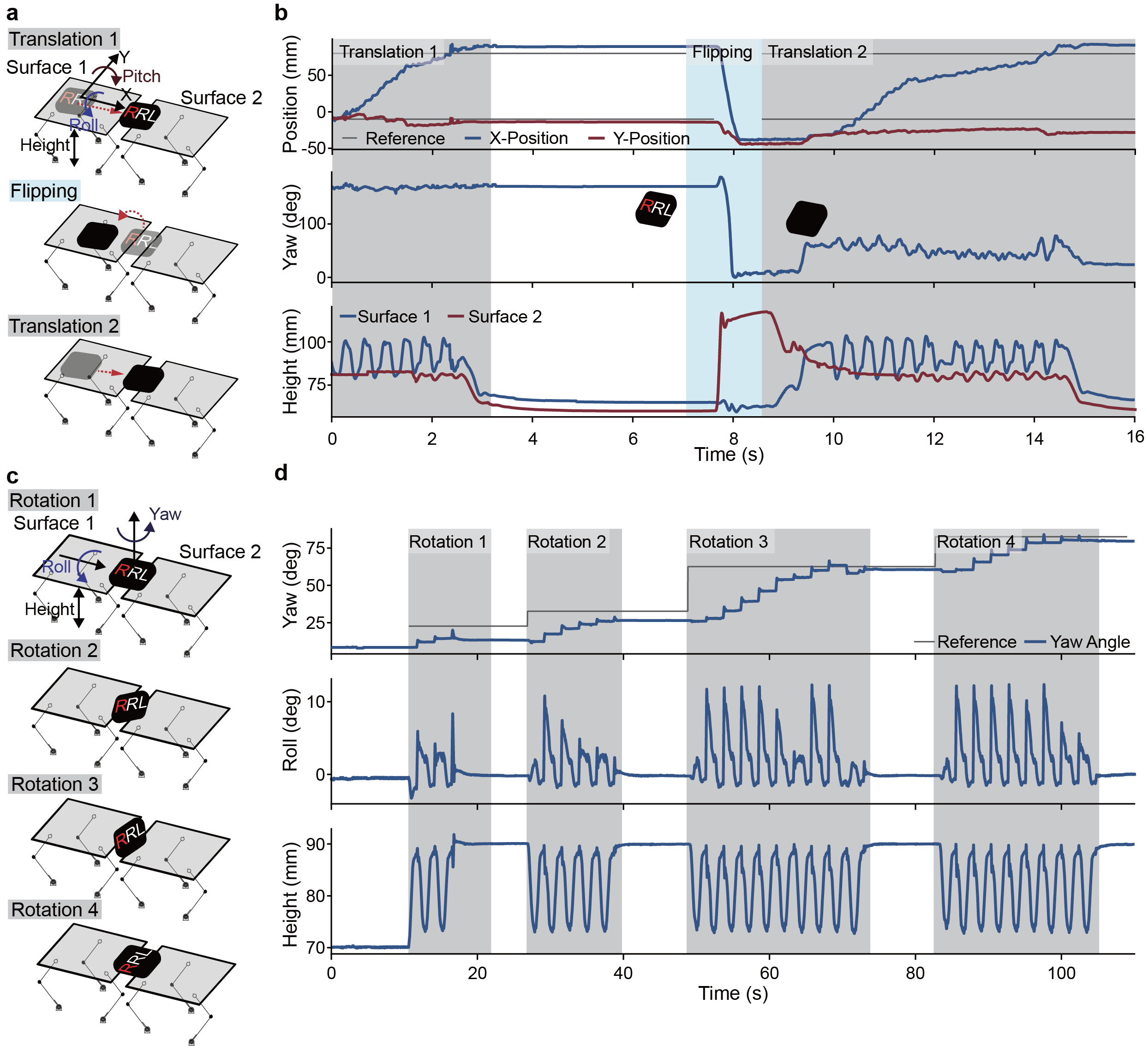}
     \caption{\textbf{Validation of translation, flipping, and rotation strategies for rigid objects.} (\textbf{a}) In translation and flipping experiments, the object (measuring $8  \times 8 \times  3\ cm^3$ and weighing $110g$) translates from an initial point on Surface 1 to between two surfaces and then flips. After flipping, the object's position shifts towards Surface 1, and it is translated again between two surfaces. (\textbf{b}), the position of the object in the surface frame is depicted over time. A noticeable $180\degree$ transition in the object’s yaw orientation in the global frame is attributed to the flipping motion. Additionally, changes in the height of Surface 1 and Surface 2 during manipulation are shown.(\textbf{c}) The schematic shows the rotation experiment setup with the object positioned between the two surfaces. The experiment sets four reference angles of $20\degree$, $30\degree$, $60\degree$, and $80\degree$ and aims to rotate the object to the target orientation. (\textbf{d}) The reference angle and the object's actual orientation are compared. It also highlights the dynamic adjustments in Surface 2's roll and height during operation.}
     \label{fig:result_of_rotation}
\end{figure*}

We tested this approach using a rectangular cuboid measuring $8  \times 8 \times  3\ cm^3$ and weighing 110 g. The object was first translated from a random starting position on Surface 1 to the intersection of two surfaces  (see Fig \ref{fig:movie}a). Knowing the object's dimensions, we designed the target position so that the object makes contact with Surface 2 while its center of gravity remains on Surface 1. This setup facilitated the flipping maneuver. After the flip, we repositioned the object between the surfaces and repeated these operations to validate both translation and flipping strategies. As shown in Fig. \ref{fig:result_of_rotation}a-b and Fig. \ref{fig:movie}a, the object was initially moved into position, flipped 180° around its Y-axis, and then returned to an intermediate location between the surfaces. Fig. \ref{fig:result_of_rotation}b tracks the object's trajectory, confirming its successful realignment. During the flipping phase, rapid height and orientation adjustments of Surface 2 briefly impacted the object, completing the flip and producing a distinct yaw change at approximately 8 seconds. Overall, these results demonstrate that our surface-based manipulation method efficiently handles complex flipping tasks, expanding the capabilities of surface-based manipulation strategies.

%TC:ignore

%TC:endignore

\subsection{Rotation of Rigid Object}

In assembly lines or food production processes, objects often need to be aligned to specific angles before packaging or further handling \cite{chen2023visual}. To demonstrate the surface-based manipulation approach for reorienting objects, we conducted an experiment to rotate a rigid object within the surfaces around the Z-axis from its initial orientation to various target orientations. Fig. \ref{fig:result_of_rotation}c presents a schematic of our experimental setup that utilizes a square prism as the object of manipulation. Building on the success of the transfer strategy validated in the previous experiment, we assumed the object's initial position to be centered between two surfaces, with its center of gravity resting on one surface, and then adjusted the other surface’s roll and height to achieve a series of target angles. In Fig. \ref{fig:result_of_rotation}d, the actual yaw angles of the object and its alignment with the set target angles are shown. A new reference angle is assigned after the object reaches a reference position. Fig. \ref{fig:result_of_rotation}d also displays the roll angle and height adjustments of Surface 1. The variation in Surface 1's roll angle tends to decrease as the discrepancy between the object's current angle and the reference angle reduces. The maximum surface's roll angle per cycle, controlled by a low-level PID controller, is capped at $12\degree$. During the experiments, we achieved an average rotational velocity of $1.16^\degree/s$. Following each interaction with the object through roll movement, the surface lowers in altitude and adjusts to an offset angle, preparing for the next interaction cycle. Through this experiment, we successfully validated the feasibility of our rotation strategy and demonstrated a novel, nonprehensile approach to reorientation.
 
\subsection{Shape Manipulation of Deformable Object}

Manipulating deformable objects is essential in various fields \cite{doi:10.1177/0278364918779698}, including manufacturing \cite{NGUYEN2021379}, the food handling \cite{satici2022coordinate}, healthcare \cite{schmidgall2024general}, and elderly care \cite{doi:10.1126/scirobotics.abm6010}. However, automating the handling of deformable objects remains a significant challenge due to the highly nonlinear character of these materials, which makes accurate modeling extremely difficult \cite{doi:10.1126/science.aat8414, yin2021}. Additionally, using grasping-based methods can cause further unpredictable deformations during manipulation \cite{doi:10.1126/scirobotics.abm6010}. Surface-based manipulation simplifies perception and control requirements because it reduces the complexity associated with modeling contact dynamics and grasp stability. In this experiment, we combined the previously introduced strategies to fold a strip of putty multiple times, relying only on measurements of the object's length and geometric center, showcasing the endless possibilities of surface-based methods. Extending this strategy to additional or larger modules could potentially enable more complex tasks involving deformable objects, such as folding clothes or kneading dough.

\begin{figure*}[t]
      \centering
      \includegraphics[width=\textwidth]{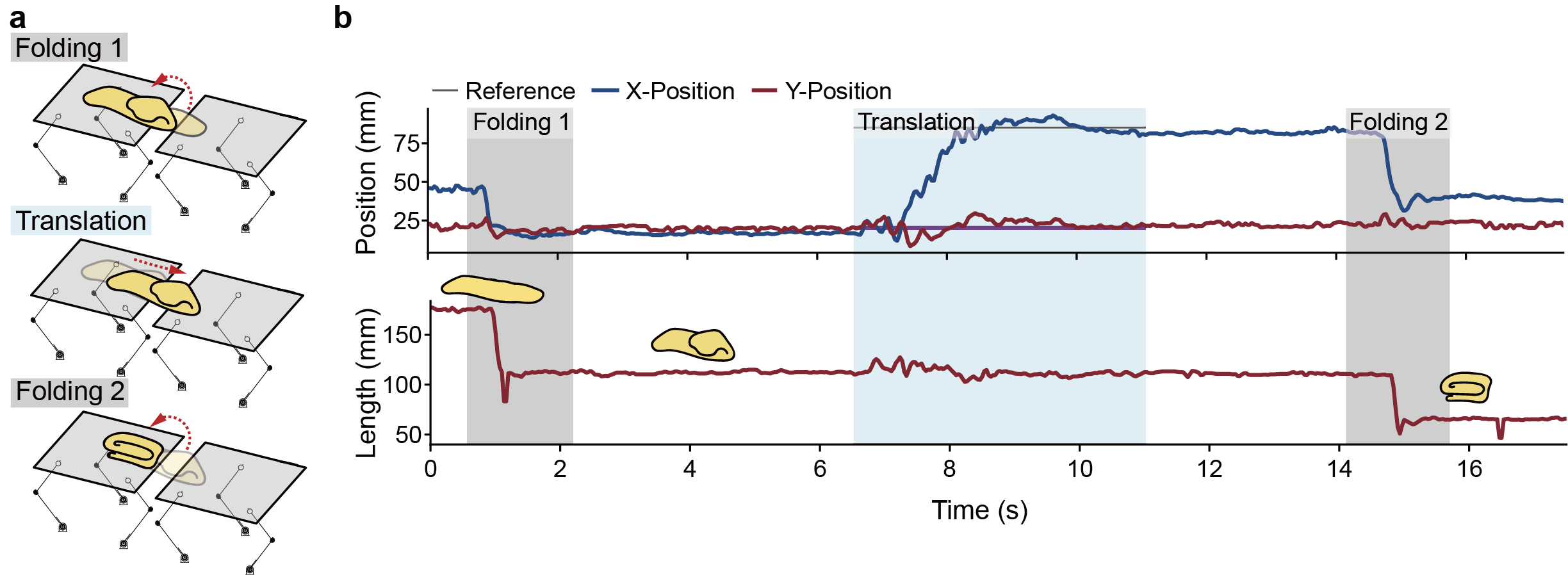}
      \caption{\textbf{Shape manipulation of free-form objects.} (\textbf{a}) Method for folding a long strip of a deformable object. The process starts with a flip for Folding 1, followed by a translation step to adjust the object for contact with both surfaces and then followed by Folding 2. (\textbf{b}) The object's XY-position and length changes throughout the experiment, with noticeable shifts at second 1 and second 15. These shifts occur because folding alters the object's geometry, which affects both its geometric center and the camera's tracking center.}
      \label{fig:result_of_softf}
\end{figure*}

In Fig. \ref{fig:movie}b and Fig. \ref{fig:result_of_softf}a, we show the process of folding a strip deformable object. In the experiment, we track the length and geometric center of the deformable object strip using a camera and apply a combination of the previously proposed transfer and flipping strategies to achieve the folding of the deformable object. At the initial stage of the experiment, we positioned the object between two surfaces, upon which we implemented our flipping strategy. However, as the object being manipulated is a soft continuum this time, only a part folds rather than undergoing a complete flip. The folding alters the object's length, and since we are tracking its geometric center, its position coordinates also change instantaneously. The position coordinates of the tracked object, as shown in Fig. \ref{fig:result_of_softf}b, indicate that the first fold occurs at second 1. The variations in object length are displayed in Fig. \ref{fig:result_of_softf}b. Subsequently, we implemented the translation strategy to position the object between the two surfaces for the next folding phase, as shown during the transfer stage in Fig. \ref{fig:result_of_softf}b. Given our knowledge of the object's length and the aim to fold approximately one-third of it in each attempt, we were able to determine the target reference positions based on the anticipated length of the fold. This approach allowed for precise control of the folding process. We executed the second folding motion once the object reached the reference position. Finally, through the integration of these operations, we successfully transformed a deformable object strip into a rolled-up configuration and changed its length. This demonstrates the versatility of surface-based manipulation and their potential applications across various fields. While our method successfully folded a low-stiffness deformable object (Young's modulus approximately between 100 MPa and 200 MPa \cite{ashraf2022mechanical}), handling more stiff materials presents additional challenges. Increased stiffness can prevent full folding and introduce greater reaction forces at contact points and lead to unintended object displacement. Future adaptations could involve implementing controlled force application, incorporating external constraints to aid bending, or estimating stiffness to refine control strategies.

\section{Discussion}\label{sec12}

In this work, we presented a method to overcome the challenge of manipulating soft and deformable objects utilizing modular robotic surfaces and successfully demonstrated three fundamental operations: translation, rotating, and flipping objects on the surface. Unlike traditional grasping methods that require selecting specific grasp points and struggle with large, low-friction, or deformable objects, our surface-based approach inherently extends to objects of different shapes, sizes, and stiffness levels. The hierarchical control structure enables the proposed strategies to be adapted to other hardware setups by using a simple plane as the end-effector and ensuring the hardware has roll, pitch, and height DoF. For instance, two conventional 6 DoF robotic arms equipped with plane end-effectors can apply the strategies presented in this study.

{Our work presents a new approach to manipulation, particularly for handling deformable, soft, and fragile objects. This has potential applications in industries such as food processing, where packaging and handling items like chicken or fish, which are typically slippery and easily damaged, pose challenges for traditional grasping methods. Using vision-based closed loop feedback, we achieved robust control by tracking the geometric center position and orientation of objects with known shapes and sizes. However, this approach is limited in dynamic environments, especially when object shape, friction, and stiffness are unknown. Although friction is not explicitly modeled, we rely on posture-based closed loop control and low-frequency vibrations, which may be ineffective for high-friction (static coefficient around 0.8) or delicate objects. Future work will focus on integrating multi-modal perception to better estimate object properties and adopting learning-based control strategies that adapt to varying conditions and object types. We also plan to explore active friction modulation methods and low-friction materials for improved friction response. In terms of scalability, our current setup uses two robotic modules with 150×150 $mm^2$ top surfaces. We will investigate both upscaling and miniaturization—larger modules (up to 500×500 $mm^2$) for heavy-load tasks, requiring stronger actuators such as high-torque motors or hydraulics, and smaller modules (5×5–10×10 $mm^2$) for integration into robotic palms with actuation method and material selection adapted to the scale. Expanding to four or six modules will enlarge the workspace, enable more deformation modes, and support simultaneous manipulation of multiple objects, increasing efficiency for tasks like automated food handling. Finally, the dynamic manipulation principles demonstrated in prior works (e.g., tossing or flipping by momentum \cite{R2cite2}) could also be integrated into our system, further extending its capabilities.}

% \textit{Future work will also explore how scaling module size and number can expand this approach’s applicability. Larger modules can handle heavier objects but require stronger actuators and more robust structures. Smaller modules suit compact systems (e.g., robotic palms) but demand careful miniaturization and flexible materials. Additionally, adding more modules will enable multi-object manipulation, enhances dexterity, and expands the workspace.}

\section{Methods}\label{sec11}

This section describes the system’s design and the perception and control methods for manipulating non-spherical objects on reconfigurable surfaces. We employ an origami-inspired foldable three-DoF parallel mechanism to achieve compactness while providing flexible joint actuation and a large workspace. For closed-loop control, we developed vision-based method to track the object’s pose in real time. Additionally, we proposed a hierarchical control strategy that outputs commands directly at the end-effector pose level rather than at the joint level for coordination among multiple modules, which makes the control strategy applicable to varying numbers of modules and diverse geometric configurations.

\subsection{Surface-Based Integrated Design}

The reconfigurable surface implemented in this project comprises two identical robotic modules. Each module has five main components (see Supplementary Fig. 1a). The surface platform is manufactured from an acrylic sheet, serving as the interface for contacting and manipulating objects via reconfigurability. The origami legs consist of a polyimide layer sandwiched between two FR4 layers, adhered together using two adhesive layers activated through a heat press application. The origami design derives from the fundamental waterbomb origami fold that can enable motion across three DoF, similar to the structures and their functionalities in the previous works \cite{Salerno2020, mintchev2019portable, Mete2021, robertson2021soft, baines2023multi}. The legs facilitate the system's reconfigurability, with each leg actuated by direct-driven BLDC motors (maxon EC-i 40) linked with 3D-printed PLA connectors. Finally, the base platform fabricated from an acrylic sheet secures the module onto the surface where the modules are fixed. Each motor, allocated to actuate the robotic modules, operates with a $36 V$ nominal voltage and can generate a nominal torque of up to $207 mN.m$. The STSPIN32F0A module (by STMicroelectronics) performs low-level control of the motors and manages the required current flow from the power supply. A main controller unit (Teensy MicroMod by PJRC and SparkFun) functions to handle high-level control operations and coordinates communication between these two boards. This communication utilizes the MODBUS RS-485 protocol and the Teensy controller's TTL output. MAX485 module converts the main controller's TTL output to the MODBUS RS-485 protocol, creating a coherent and reliable communication channel between the low-level actuation control and high-level functions.

\subsection{Integrated Perception Systems for Real-Time Object States Tracking on Surfaces}

In order to obtain accurate position and orientation data for both the object positioned on the reconfigurable surface and each individual module's surface, we employed a perception system based on computer vision. For objects characterized as nearly-rigid, we utilized a motion capture setup, equipped with six high-field-of-view and high-precision cameras (Vicon Vero). These objects carried deliberately placed markers to perform accurate tracking. The data obtained from the motion capture system was then transferred to the closed-loop control setup through an I2C communication protocol that was set to operate at a frequency of $100 Hz$. We adopted an alternative approach for the objects having deformable materials and non-rigid behavior where marker attachment is not possible. A camera-based custom image processing setup was established, employing a Logitech C270 webcam and OpenCV. This setup was utilized to detect the object's geometrical center and its boundaries. These methods enabled us to obtain the position feedback for manipulating both deformable and non-deformable objects on the reconfigurable surface.
 
\subsection{Control System Overview and Implementation}

The overall control of the developed system includes five main components (Fig. \ref{fig:control_scheme}). Manipulation strategy transitions are managed by predefined task objectives. This system employs a state machine that dynamically alters between three modes: translation, rotation, and flipping. The position and the orientation of the object are obtained by the real-time sensor data from the vision system, guiding the dynamic generation of the surfaces' roll, pitch, and height references in accordance with the appropriate manipulation strategy. Using the inverse kinematics algorithm explained in Supplementary Notes 1, the desired joint angles are computed. These reference values are then input into a PID-based servo motor controller, ensuring precise trajectory tracking of the joint commands. In the control scheme for the BLDC motor, as illustrated in Fig. \ref{fig:control_scheme}c, a cascade control structure is implemented, comprising both a feedforward component and a current loop within the position loop.

\subsection{High-Level Controller for Three Different Surface-Based Manipulation Strategies: Translation, Rotation, Flipping}
\label{sec:Strategies}

\subsection*{Translation}Surface-based manipulation enables moving an object from any initial position to a specified target. Our translation strategy focuses on objects that fit entirely on a single surface. As shown in Fig. \ref{fig:control_scheme}, given the object's current center of gravity (CoG) position, the goal is to move its centroid to the target. Here, $X_{error}$ and $Y_{error}$ are the positional errors along the local X- and Y-axes of the surface.

We define the surface orientation with XYZ-Euler angles and set $yaw=0$ due to the Canfield structure’s directional constraint. Algorithm \ref{al:Translation algorithm} outlines the complete procedure. Each control cycle computes pitch and roll references based on $X_{error}$ and $Y_{error}$, with a PID controller determining the exact angles (Fig. \ref{fig:control_scheme}b and Eq. \ref{eq:trans_roll_pitch}):

\begin{algorithm}
\caption{Translation Strategy}
\label{al:Translation algorithm}
\begin{algorithmic}[1]
\Procedure{Translation}{$ X_{ref}, Y _{ref}, X_{cur}, Y_{cur},\epsilon$}

\State {\textcolor{mydarkgreen}{$\#\ $Line 3-4: compute the position error between the target and the object’s current position.}}
\State $X_{error} =  X_{ref}- X_{pos}$  
\State $Y_{error} =  Y_{ref}- Y_{pos}$ 
\While{$ \left| X_{error} \right| + \left| Y_{error} \right|  \ge \epsilon$}
\State {\textcolor{mydarkgreen}{$\#\ $Line 7-10: if the positional error exceeds $\epsilon$, calculate the reference pose for the surface.}}
\State $\theta_{surface\_ roll}=\textrm{PID}(Y_{error})$ \Comment{Equation \ref{eq:trans_roll_pitch}}
\State $\theta_{surface\_ pitch}=\textrm{PID}(X_{error})$ \Comment{Equation \ref{eq:trans_roll_pitch}}
\State $H=A \sin(\omega t)+ H_{init}$ \Comment{Equation \ref{eq:surfaceheight}}
\State  $Surface \gets \left[\theta_{Surface1\_roll},\theta_{surface\_ pitch},H \right]$     
\EndWhile

\EndProcedure
\end{algorithmic}
\end{algorithm}

\begin{equation}
\begin{split}
    Ref_{roll} &=K_PY_{error}(t)+K_i\int_{0}^{t} Y_{error}(\tau) d\tau  +K_d\frac{d}{dt}Y_{error},\\
    Ref_{pitch} &=K_PX_{error}(t)+K_i\int_{0}^{t} X_{error}(\tau) d\tau  +K_d\frac{d}{dt}X_{error}.
\end{split}
    \label{eq:trans_roll_pitch}
\end{equation}

In low-friction scenarios, where the friction coefficient $\mu$ is smaller than $tan(\psi)$ (the tangent of the surface's normal vector polar angle $\psi$), the control approach resembles a classic ball balancing table. Here, adjusting the surface’s inclination smoothly guides the object to its target. When the friction coefficient between two objects exceeds 0.33, inclination alone may not suffice. To overcome this, we modulate the surface’s height as well as its orientation (Fig. \ref{fig:control_scheme}b), introducing vertical oscillations that reduce frictional resistance and apply periodic impulses to move the object. We define:

\begin{equation} 
H = A \sin(\omega t) + H_{init},
\label{eq:surfaceheight}
\end{equation} 

where $t$ is the elapsed time and $\omega$ is the oscillation frequency, adjustable to suit different objects. The elevation of the surface also affects its maximum achievable tilt angle (see Supplementary Fig. 2). To maintain a large range of motion, here the vertical displacement amplitude spans from 8 cm to 10 cm. Friction affects manipulation by influencing the required oscillation amplitude and movement speed, creating a trade-off between operational speed and the safe handling of delicate objects.

Using Euler angles, the rotation matrix of the top surface can be calculated in Equation \ref{eq:rotationmtrix}, where $\phi$ is the reference roll angle, and $\theta$ is the reference pitch angle:

\begin{equation}
    C_{WR}= 
    \begin{bmatrix}
        &\cos(\theta)\ &0\ &sin(\theta) \\ 
    &\sin(\phi)\sin(\theta)\ &\cos(\phi)\ &-\cos(\theta)\sin(\phi)\\
    &-\cos(\phi)\sin(\theta)\ &\sin(\phi)\ &\cos(\phi)\cos(\theta)\\
    \end{bmatrix}.
    \label{eq:rotationmtrix}
\end{equation}

Neglecting vertical translation, the top surface’s normal vector is:

\begin{equation}
    N_{O_R} = C_{WR}*[0,0,1]^T = [\sin(\theta), -\cos(\theta)\sin(\phi), \cos(\phi)\cos(\theta)]^T.
\end{equation}

From $N_{O_R}$, we derive the polar and azimuthal angles of the surface:

\begin{equation}
\begin{split}
    \delta &=arctan(\frac{N_{O_R}(y)}{N_{O_R}(x)})=\arctan(\frac{-\cos(\theta)\sin(\phi)}{\sin(\theta)}),\\
    \psi   &=\arccos(N_{O_R}(z))=\arccos(\cos(\phi)\cos(\theta)).
\end{split}
    \label{eq:delta_psi}
\end{equation}

Once we determine the required orientation and height parameters $(\delta,\psi,H)$ for the Canfield mechanism’s end-effector, we apply inverse kinematics (Supplementary Notes 1) to compute the corresponding joint angles.

\begin{algorithm}
\caption{Rotation Strategy}
\label{al:rotation algorithm}
\begin{algorithmic}[1]
\Procedure{Rotation}{$\theta _{ref
}, \theta _{yaw}, \epsilon$}
\State {\textcolor{mydarkgreen}{$\#\ $Line 3-4: initialize the surface.}}
\State $Surface_1 \gets \left[0,0,H_{init} \right]$ 
\State $Surface_2 \gets \left[0,0,H_{init} \right]$
\State {\textcolor{mydarkgreen}{$\#\ $Line 6-21: repeat when the angular error between the object’s current angle and target angle exceeds $\epsilon$. This constitutes one cycle.}}
\While{$ \left|\theta _{ref
}- \theta _{yaw} \right|  \ge \epsilon$} 

\State $\theta_{surface2\_ roll}=\textrm{PID}(\theta_{ref} - \theta_{yaw})$  \Comment{Equation \ref{eq:rotation_roll}}
\If {$ \theta_{surface2\_ roll} \ge 0$}
    \State $\theta_{Surface1\_roll}=-\theta_\gamma$
\Else
    \State $\theta_{Surface1\_roll}=\theta_\gamma$
\EndIf
\State {\textcolor{mydarkgreen}{$ \#\ $rotate the surface to the reference pose to induce the object’s rotation.}}
\State  $Surface_1 \gets \left[\theta_{Surface1\_roll},-\theta_{\beta},H_{init} \right]$     
\State  $Surface_2 \gets \left[\theta_{Surface2\_roll},-\theta_{\beta},H_{init} \right]$
\State $delay\ 500ms$
\State {\textcolor{mydarkgreen}{$\#\ $reset the surface after one cycle to establish new contact points in next cycle.}}
\State $Surface_1 \gets \left[0,0,H_{init} \right]$    
\State $Surface_2 \gets \left[0,\theta_{\beta},H_{low} \right]$
\State $Surface_2 \gets \left[0,\theta_{\beta} ,H_{init} \right]$
\EndWhile
\State {\textcolor{mydarkgreen}{$\#\ $Line 23-24: reset the surface once the object reaches the reference angle to prepare for the next task.}}
\State $Surface_1 \gets \left[0,0,H_{init} \right]$ 
\State $Surface_2 \gets \left[0,0,H_{init} \right]$
\EndProcedure
\end{algorithmic}
\end{algorithm}

\subsection*{Rotation}
We achieve surface-based object rotation by repeatedly establishing and breaking contact between the object and multiple surfaces. As the object interacts with these surfaces, their differing force directions enable complex rotational maneuvers (see Fig. \ref{fig:control_scheme}b). We define the object’s current yaw angle as $\theta_{yaw}$ and the target angle as $\theta_{ref}$.

Building on the XYZ-Euler angle coordinate, we modulate Surface 2’s roll angle to induce torque on the object (Fig. \ref{fig:control_scheme}b), as determined by a PID controller (Eq. \ref{eq:rotation_roll}):

\begin{equation}
    \theta_{surface2 \_roll}=K_P*\theta_{error}(t)+K_i\int_{0}^{t} \theta_{error}(\tau) d\tau  +K_d\frac{d}{dt}\theta_{error}.
    \label{eq:rotation_roll}
\end{equation}

Algorithm \ref{al:rotation algorithm} details the rotation procedure. Each cycle adjusts surface angles to exert torque, then returns them to a neutral state, ensuring fresh contact points for subsequent cycles. During rotation, we adjust the pitch angle of Surface 1 to incline toward the center, against object displacement along the X-axis. Additionally, the roll of Surface 1 is also adjusted by a fixed angle $\theta_\gamma$ in correspondence with the object's rotation direction, further facilitating the object's rotation. In cases where the surface reaches its maximum roll angle before achieving the target object orientation, we realign the surfaces horizontally and then raise Surface 2 back to its initial height, providing a renewed contact interface (Algorithm \ref{al:rotation algorithm} line 19-20). If the object still does not reach the desired angle, the algorithm repeats these steps (Algorithm \ref{al:rotation algorithm} line 6-21) until convergence.

\begin{algorithm}
\caption{Flipping Strategy}
\label{al:flipping algorithm}
\begin{algorithmic}[1]
\Procedure{Flipping}{}
 
\State {\textcolor{mydarkgreen}{$\#\ $Line 3-5: based on the object's size, transport it from the initial position to a flipping position between the two surfaces.}}
\If {$ \|(X_{ref},Y_{ref})-(x,y)\| > \epsilon $} 
   \State $Surface_1 \gets \textrm{translation}\left(X_{ref},Y_{ref}\right)$ \Comment{Algorithm \ref{al:Translation algorithm}}
   \State $Surface_2 \gets \left[0,-\theta_{\beta},H_{init}-1\right]$  
\Else
\State {\textcolor{mydarkgreen}{$\#\ $Line 8-9: Lower the surface to initialize the flipping task.}}

    \State $Surface_1 \gets \left[0,0,H_{low} \right]$
    \State $Surface_2 \gets \left[0,\theta_{\beta},H_{low} \right]$
    \State $delay\ 5000ms$
    \State $Surface_1 \gets \left[0,0,H_{low} \right]$   
    
    \State {\textcolor{mydarkgreen}{$\#\ $Quickly raise Surface 2 to generate an impact, flipping the object.}}
    \State $Surface_2 \gets \left[0,0,H_{high} \right]\ in\  100ms$  
    \State $delay\ 1000ms$  
    \State {\textcolor{mydarkgreen}{$\#\ $Line 16-17: after flipping, reset the surface to prepare for the next task.}}
    \State $Surface_1 \gets \left[0,0, H_{init}\right]$ 
    \State $Surface_2 \gets \left[0,0,H_{init} \right]$
\EndIf
\EndProcedure
\end{algorithmic}
\end{algorithm}

\subsection*{Flipping} Flipping an object by 180° requires surface delivering substantial impulses within a brief timescale. As shown in Fig. \ref{fig:control_scheme}b, our approach begins with the object positioned in contact with two surfaces, with its CoG falling into Surface 1. To initiate a flipping around a specific contact point, Surface 2 undergoes a rapid elevation, exerting a continuous force onto the object's one edge. To enhance this rotational trajectory, Surface 2’s initial ascent deviates from a horizontal orientation by adopting a fixed angular tilt relative to the Y-axis as detailed in line 9 of Algorithm \ref{al:flipping algorithm}.

As Surface 2 pushes upward, the object’s angular velocity increases. Eventually, Surface 2 reaches its maximum elevation and loses contact, leaving the object to continue rotating under gravity alone. For a successful flip, the object’s angular velocity must remain positive when the line connecting its CoG and the contact point becomes perpendicular to the surface, allowing gravity to complete the rotation.

Let $\omega_{t1}$ be the angular velocity of the object when it separates from Surface 2. The kinetic energy at this point must be equal to or surpass its gravitational potential energy for the flipping action to succeed. Mathematically, this is expressed as follows: 

\begin{equation}
\begin{split}
0-\frac{1}{2}J\omega_{t1}^2 &\geq \frac{L}{2}mg(\sin(\theta)-1),\\
\omega_{t1\_min} &= \sqrt{\frac{mgL(1-\sin(\theta))}{J}}.
\end{split}
\label{eq:phase2}
\end{equation}

Where $\theta$ is the angle between the target object and the horizontal when Surface 2 detaches, $J$ represents the target object's rotational inertia on this axis, and $L$ is the object's cross-sectional length.

The main strategy is centered on optimizing the linear velocity of Surface 2. Adjusting Surface 2 to reach this critical velocity level can provide the object with enough momentum to overcome gravitational forces:

\begin{equation}
   V_{min\_vel
   \_required} = L*\omega_{t1\_min}.
\end{equation}

Another approach is to maximize Surface 2’s acceleration, pushing it to the highest possible speed. This strategy delivers maximum momentum to the object, reducing uncertainties and increasing the likelihood of a successful flip.

\backmatter
%TC:ignore

\section*{Declarations}
\subsection*{Data availability}
All data needed to evaluate the conclusions in the paper are present in the paper or the Supplementary Materials.

\subsection*{Acknowledgements}

The authors thank Alexander Schüßler of Reconfigurable Robotics Lab, EPFL for his contributions to using computer vision to identify the position of deformable objects. This research has received funding from the European Union’s Horizon Europe research and innovation programme under grant agreement no: 101069536 (MOZART project).

\subsection*{Author contribution}
\begin{itemize}
    \item Conceptualization ZW, SD, FZ, and JP
    \item Funding acquisition JP
    \item Investigation ZW, SD
    \item Methodology ZW, SD, and JP
    \item Software ZW and SD
    \item Supervision JP
    \item Visualization ZW, SD, and FZ
    \item Writing - original draft ZW and SD
    \item Writing - review \& editing ZW, SD, FZ, and JP
\end{itemize}
\subsection*{Competing Interests}
The authors declare no competing interests. 
\begin{appendices}

\end{appendices}

\bibliography{SensorizedTile}

%TC:endignore
\newpage

%TC:ignore

\end{document}